\documentclass[]{interact}

\usepackage{algcompatible}
\usepackage{algorithm}
\usepackage{siunitx}
\usepackage{multirow}
\usepackage{hyperref}
\hypersetup{
	colorlinks=true,
	citecolor=blue,
}

\usepackage{epstopdf}
\usepackage{subfigure}

\usepackage{natbib}
\bibpunct[, ]{(}{)}{;}{a}{}{,}
\renewcommand\bibfont{\fontsize{10}{12}\selectfont}

\theoremstyle{plain}

\theoremstyle{definition}

\theoremstyle{remark}

\def\mathbf#1{\boldsymbol{#1}}

\begin{document}

\articletype{ARTICLE TEMPLATE}

\title{Deep Subspace Learning for Surface Anomaly Classification Based on 3D Point Cloud Data}

\author{
\name{Xuanming~Cao\textsuperscript{a*}, Chengyu Tao\textsuperscript{b*} and Juan Du\textsuperscript{a,b}\thanks{CONTACT Juan Du. Email: juandu@ust.hk; * Equal contribution}}
\affil{\textsuperscript{a}Smart Manufacturing Thrust, Systems Hub, The Hong Kong University of Science and Technology (Guangzhou), Guangzhou, China; \textsuperscript{b}Interdisciplinary Programs Office, The Hong Kong University of Science and Technology, Hong Kong SAR, China}
}

\maketitle

\begin{abstract}
Surface anomaly classification is critical for manufacturing system fault diagnosis and quality control. 
However, the following challenges always hinder accurate anomaly classification in practice: (\romannumeral1) Anomaly patterns exhibit intra-class variation and inter-class similarity, presenting challenges in the accurate classification of each sample. (\romannumeral2) Despite the predefined classes, new types of anomalies can occur during production that require to be detected accurately.
(\romannumeral3) Anomalous data is rare in manufacturing processes, leading to limited data for model learning. To tackle the above challenges simultaneously, this paper proposes a novel deep subspace learning-based 3D anomaly classification model. Specifically, starting from a lightweight encoder to extract the latent representations, we model each class as a subspace to account for the intra-class variation, while promoting distinct subspaces of different classes to tackle the inter-class similarity. Moreover, the explicit modeling of subspaces offers the capability to detect out-of-distribution samples, i.e., new types of anomalies, and the regularization effect with much fewer learnable parameters of our proposed subspace classifier, compared to the popular Multi-Layer Perceptions (MLPs). 
Extensive numerical experiments demonstrate our method achieves better anomaly classification results than benchmark methods, and can effectively identify the new types of anomalies.
\end{abstract}


\section{Introduction}

\label{introduction}
Surface anomalies are inevitable in complex manufacturing processes, such as additive manufacturing and forming, which can significantly compromise their structural integrity and lead to product failure \citep{tao2023anomaly}. Different types of anomalies may correspond to distinct root causes. Subsequently, accurate anomaly identification is vital for root cause diagnosis, essential to maintaining production operations and improving manufacturing systems. 

With the advance of 3D scanning technology, 3D point cloud data has garnered increasing attention in various industrial applications \citep{colosimo2022complex, shi2019identifying}, such as dimensional inspection \citep{molleda2013improved} and shape variation analysis \citep{bui2022analyzing}. Moreover, 3D point cloud data provides richer geometrical information about surface anomalies \citep{tao2025pointsgrade} compared to the 2D image data, which are valuable for anomaly classification. Therefore, it is crucial to develop an accurate anomaly classification model based on 3D point cloud data.

\begin{figure}[]
    \centerline{\includegraphics[width = 0.6\linewidth]{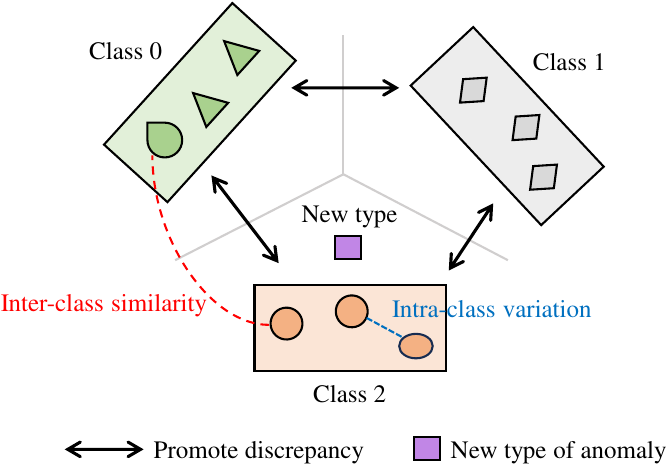}}
    \caption{The motivations of proposed deep subspace model for anomaly classification. (\romannumeral1) Different samples of the same class can be located widely in the same subspace, which allows for intra-class variation. (\romannumeral2) The subspaces are promoted to be discrepant so that similar samples across different classes can be distinguished. (\romannumeral3) The new type of anomaly is far away from all established subspaces and thus able to be detected.}
    \label{motivation}
\end{figure}

In the literature, current research on 3D anomaly classification is still limited. To the best of our knowledge, the following literature has been identified. For example, \cite{du2022tensor} defined several descriptive statistics for anomalies on steel surfaces, which were further classified by a support vector machine. Furthermore, various deep neural networks including graph neural network \citep{wang2023mvgcn} and attention network \citep{hu2022efficient}, originally developed for general 3D object classification, were introduced to facilitate an end-to-end anomaly classification, with applications to the data from concrete specimens \citep{wang2023mvgcn} and solder joints \citep{hu2022efficient}. A comprehensive review is presented in Section \ref{related}.

However, additional challenges prevalent in industrial applications remain inadequately addressed by existing studies, thereby impeding precise anomaly classification:
\begin{itemize}
    \item Due to the complexity and randomness of real-world manufacturing systems, anomalies always exhibit complex patterns. Specifically, anomalies within the same class display varied shapes and sizes, representing \textit{intra-class variation}, while those across different classes may exhibit similarities, denoted as \textit{inter-class similarity}. This characteristic complicates the effective learning of decision boundaries between different classes.

    \item New types of anomalies will continuously emerge throughout production, underscoring the impossibility of collecting all potential anomalies in advance \citep{sun2023continual}. However, existing classification models, which are limited to predefined classes during training, fail to detect unseen classes that newly appear during production and mistakenly categorize them as known ones.

    \item Anomalous data is rare in manufacturing processes, thereby hard to collect sufficient anomalous samples. Lack of sufficient data may lead to overfitting in many data-driven methods, including deep learning-based methods.
    
\end{itemize}

To address the above challenges, we propose a novel deep subspace model to achieve accurate anomaly classification based on 3D point cloud data.  Firstly, we assume that each class of anomaly occupies a specific subspace. Samples of the same class can be located variably within the subspace, allowing the intra-class variation. Simultaneously, we promote distinct subspaces to enhance the discrepancy among various classes. Therefore,  similar samples from different classes can be discriminated against, alleviating the issue of inter-class similarity. Furthermore, the new types of anomalies will be out-of-distribution \citep{maboudou2024deep}, i.e., their representations are far from the learned subspaces, serving as a basis for the identification of them. We illustrate the above motivation of our deep subspace model in Fig. \ref{motivation}. Finally, modeling each class as a subspace decreases the number of parameters compared to the popular Multi-Layer Perception (MLP) classifier, offering improved generalization and regularization \citep{chen2017deep} to mitigate overfitting.

To achieve this, we adopt a lightweight encoder to transform input point clouds into discriminative representations. Subsequently, we propose a novel subspace classifier and related loss function to encourage the desired properties of encoded representations. Additionally, we also propose a practical strategy to ensure better model training.

To sum up, the main contributions of this paper are as follows:
\begin{itemize}
    \item We propose a novel unified model for accurate anomaly classification and simultaneous identification of new anomaly types.
    
    \item We innovatively model each class through subspace learning, explicitly tackling the intra-class variation and inter-class similarity of anomalies, and capable of identifying new types of anomalies and alleviating overfitting.

    \item We design the specific architecture of the subspace classifier, corresponding loss function, and a practical training strategy to enable efficient model learning for our task.
\end{itemize}

The structure of this paper is delineated as follows. Section \ref{related} reviews the literature about general 3D point cloud modeling and 3D anomaly classification. Subsequently, Section \ref{method} elucidates the development of our deep subspace learning methodology. The classification and detection efficacy of our model is empirically validated in Sections \ref{exp_num} and \ref{exp_case}. Finally, this paper is concluded in Section \ref{conclusion}.

\section{Related Work}
\label{related}

This section commences with the deep learning-based methods for general 3D point clouds, subsequently delving into specialized techniques for 3D anomaly classification.

\subsection{Deep Learning for General 3D Point Clouds}

Unlike 2D images, 3D point cloud data are irregular, permutation invariant, and have an additional dimension, making them incompatible with conventional Convolutional Neural Networks. To deal with it, PointNet \citep{qi2017pointnet} adopted a shared MLP to encode points into latent features, further aggregated into a global one for classification. To incorporate the local semantics ignored by PointNet, the extended PointNet++ \citep{qi2017pointnet++} proposed a hierarchical framework to aggregate information from local regions. Furthermore, some approaches represented point clouds as graphs that were processed by graph neural networks \citep{te2018rgcnn,wang2019dynamic}. Among them, DGCNN \citep{wang2019dynamic} dynamically updated the graph structure during learning, enhancing the local information sharing and thus improving the performance of classification and semantic segmentation. Recently, Transformers have excelled in natural language processing through self-attention mechanisms that capture global data dependencies. This architecture, applied in 3D point cloud analysis, e.g., the Point Cloud Transformer (PCT) \citep{guo2021pct}, has also achieved impressive outcomes in multiple downstream tasks.

While these methods were originally developed for general point cloud classification and segmentation tasks, they are also applicable to anomaly classification tasks. However, they fail to effectively address the challenges posed by intra-class variation and inter-class similarity, which complicate the classification process. Additionally, they are unable to handle unseen types of anomalies.

\subsection{3D Anomaly Classification Methods}

Classical machine learning-based methods consist of feature extraction and further classification. The features can be defined by the discriminated statistics of sharp points \citep{du2022tensor}, spectral graph Laplacian eigenvalues \citep{samie2017classifying}, and the histogram of deviations between the skin model and CAD model \citep{yacob2019anomaly}. In addition, the anomalies were projected into 2D images such that the characteristics including moments and Fourier descriptors were extracted \citep{madrigal2017method}. Then, various machine learning classifiers can be applied for classification, including the sparse support vector machine \citep{du2022tensor}, sparse representation \citep{samie2017classifying}, and ensemble classifiers \citep{yacob2019anomaly}. The primary limitation of these methods is that the features may lack discriminative power for complex anomaly patterns, such as intra-class variation and inter-class similarity of anomalies.

Deep learning-based methods have emerged that enable more effective feature learning.
For example, leveraging the DGCNN model \citep{wang2019dynamic}, \cite{wang2023mvgcn} introduced a multi-view framework that utilized two sub-networks to explicitly consider both anomaly and reference points, emphasizing the significance of the potentially scarce anomaly points.  Similar to the PCT model \citep{guo2021pct}, \cite{zhou2022sewer} proposed a Transformer-based network that achieved satisfactory performance on 3D sewer anomaly classification. Moreover, \cite{bolourian2023point} utilized PointNet++ to detect the surface anomalies on concrete bridge surfaces. Generally speaking, PointNet, DGCNN, and their variants are potential choices for 3D anomaly classification applications. 

However, the above deep learning-based methods do not directly target the situation with intra-class variation and inter-class similarity, and cannot detect new types of anomaly. This situation is common and important in many realistic manufacturing applications.


\section{Methodology}
\label{method}

\subsection{Overview}
\label{problem}

We denote the labeled training dataset as $ \{(\mathbf{X}_i, \mathbf{y}_i) \in \mathbb{R}^{N \times 3} \times \mathbb{R}^{C}, \allowdisplaybreaks i=1, 2, \dots, M \}$, consisting of $C$ classes and $M$ samples. Each sample comprises $N$ points positioned in 3D Cartesian space. If the $i\mathrm{th}$ sample belongs to the $c\mathrm{th}$ class, we have $\mathbf{y}_{i,c} = 1$ and 0 otherwise.

\begin{figure}[t]
    \centerline{\includegraphics[width = \linewidth]{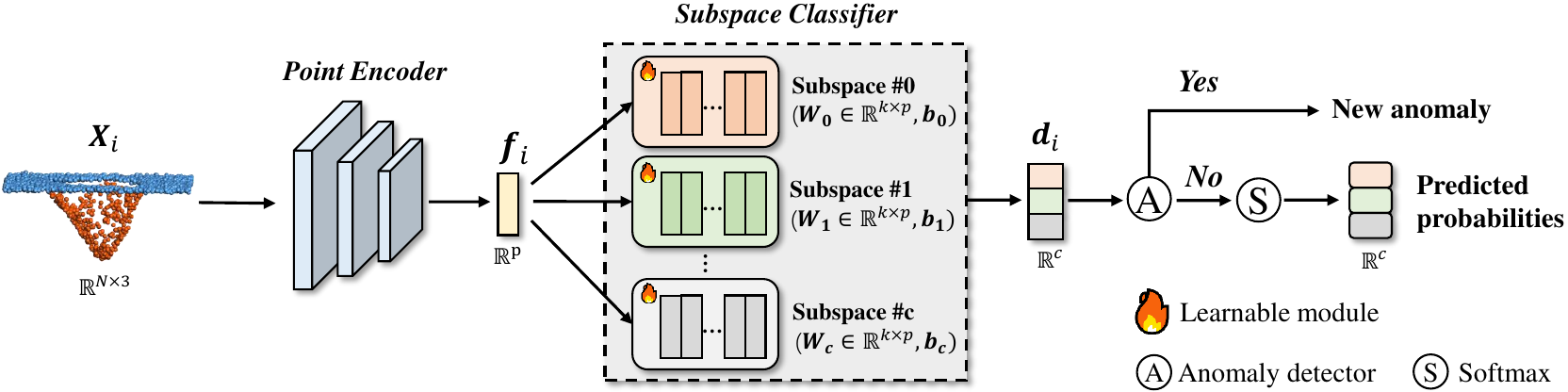}}
    \caption{Our proposed deep subspace learning framework for anomaly classification and the new type of anomaly detection.}
    \label{framework}
\end{figure}

Our proposed model contains an \textit{point encoder} for representation learning, succeeded by a \textit{subspace classifier} for detecting new types of anomaly or categorizing known anomalies. Initially, for the $i\mathrm{th}$ sample $\mathbf{X}_i$, a lightweight encoder generates its representation, $\mathbf{f}_i \in \mathbb{R}^{p}$.  This representation is subsequently analyzed by the subspace classifier, which computes the distances, $\mathbf{d}_i \in \mathbb{R}^{C}$, relative to established subspaces. A sample is identified as a new type of anomaly if all values of $\mathbf{d}_i$ are relatively large; otherwise, it is classified into a known class. The overview of our method is illustrated in Fig. \ref{framework}.

\subsection{Point Encoder}

\begin{figure}[]
    \centerline{\includegraphics[width = 0.8\linewidth]{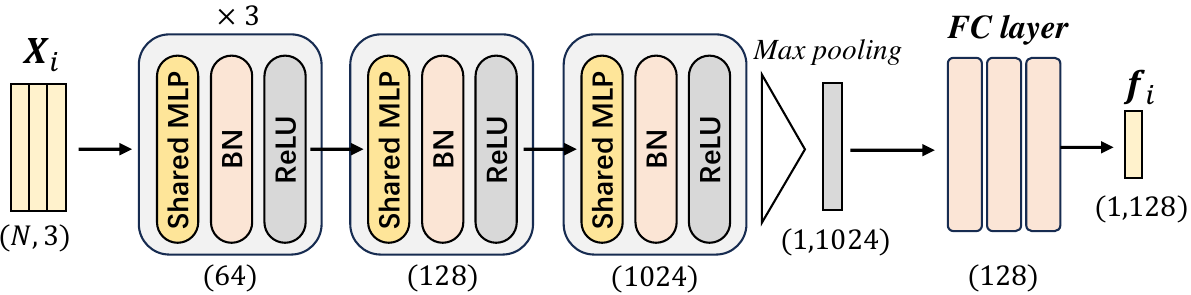}}
    \caption{The architecture of the adopted encoder when $p=128$. The shared MLP is the 1D convolution used in PointNet \citep{qi2017pointnet}, BN is batch normalization operation, and ReLU is the activation function.}
    \label{feature encoder}
\end{figure}

The encoder aims to process unstructured point cloud data, while simultaneously ensuring a limited number of parameters due to the limited availability of training data. Therefore, we adopt a lightweight PointNet-like architecture \citep{qi2017pointnet} as our encoder, as shown in Fig. \ref{feature encoder}.

The details are as follows: For the sample $\mathbf{X}_i$, we expand the number of channels from 3 to 64 using three identical encoding blocks and then map to 128 and 1024 using two more blocks. Each block comprises a shared MLP layer (i.e., 1D convolution \citep{qi2017pointnet}), a batch normalization layer, and a ReLU activation function. Finally, the 
max-pooling operation is applied to obtain a global view, i.e., a 1024-dimensional representation, which is then projected as a lower dimensional representation $\mathbf{f}_i$ by a fully connected (FC) layer.
The shared MLP and the max-pooling operation are permutation-invariant and applicable to unstructured data. The above procedure can be formulated as:
\begin{equation}
    \mathbf{f}_i=\mathcal{E}_{\boldsymbol{\theta}}(\mathbf{X}_i),
    \label{encoder}
\end{equation}
where $\mathcal{E}_{\boldsymbol{\theta}}$ is the encoder that maps $\mathbf{X}_i$ into a $p$-dimensional representation, and $\boldsymbol{\theta}$ is the set of learnable parameters.

\subsection{Subspace Classifier}

In this section, we detail the construction of subspaces and their use in anomaly classification and detection of new types of anomaly.

\subsubsection{Subspace Construction}
\label{subspace construction}

In a high-dimensional space, a subspace is spanned by a set of bases. For the $c\mathrm{th}$ class, we design the bases of subspace as a matrix $\mathbf{W}_c = \left[\boldsymbol{v}_{1,c}, \boldsymbol{v}_{2,c}, \cdots, \boldsymbol{v}_{k,c} \right] \in \mathbb{R}^{p\times k}$ and a bias $\mathbf{b}_c \in  \mathbb{R}^{p}$, where the column vector $\boldsymbol{v}_{i,c}$ is a basis of subspace $c$ and $k$ is the subspace dimension. $\{\mathbf{W}_c,\mathbf{b}_c \}$ are learnable parameters that will be estimated during training.

\begin{figure}[]
    \centerline{\includegraphics[width = 0.65\linewidth]{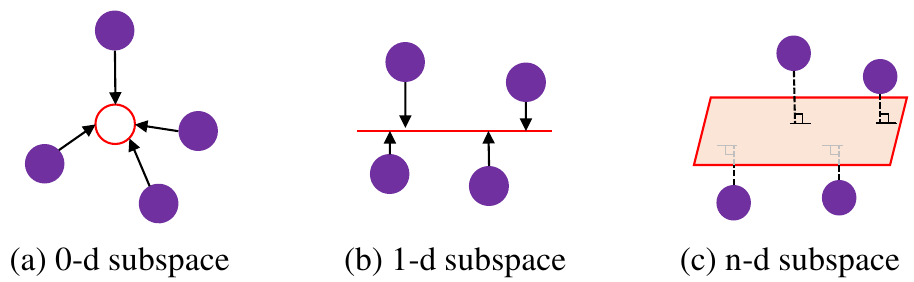}}
    \caption{Interpretation of different dimensions of subspaces. (a) 0-dimensional subspace. The subspace degrades to a singleton. (b) 1-dimensional subspace. The subspace is a line. (c) n-dimensional subspace. The subspace is a hyperplane that allows representations distributed along various directions.}
    \label{subspace dim}
\end{figure}

To understand the dimension $k$ of subspace, we provide special examples in Fig. \ref{subspace dim}. The $0$-dimensional subspace degrades into a singleton, where all representations are expected to be close. Similarly, the $1$-dimensional subspace becomes a line. Generally, increasing the dimension $k$ indicates that the subspace can capture more intra-class variation with a greater degree of freedom. However, excessively large $k$ will memorize extra noise in training data, which should be avoided.

\subsubsection{Distance Calculation}

\begin{figure}[]
    \centerline{\includegraphics[width = 0.5\linewidth]{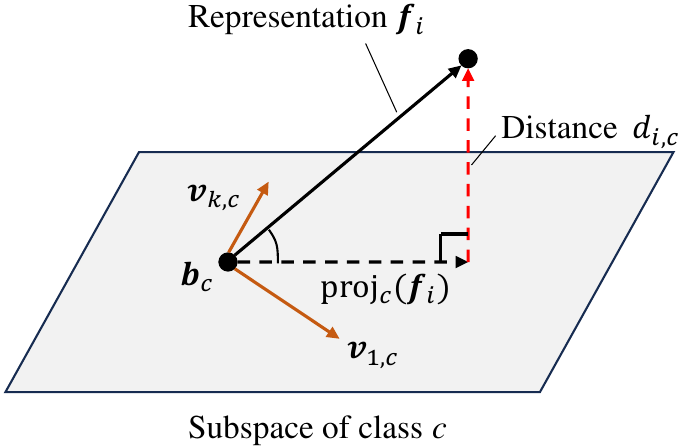}}
    \caption{Projection of representation $\mathbf{f}_i$ on the subspace of class $c$. }
    \label{subspace}
\end{figure}

Distances from subspaces are crucial for detecting new types of anomalies and classifying known ones. Specifically, the distance $d_{i,c} \in \mathbb{R}$ of $\mathbf{f}_i$ from the $c\mathrm{th}$ subspace can be calculated as follows:
\begin{gather}
    d_{i,c}=\left\|\mathbf{f}_i-\mathrm{proj}_c(\mathbf{f}_i)\right\|_2, \\
    \mathrm{proj}_c(\mathbf{f}_i) = \mathbf{W}_c(\mathbf{W}_c^{\top}\mathbf{W}_c)^{-1}\mathbf{W}_c^{\top}(\mathbf{f}_i-\mathbf{b}_c),
\label{deviation cal}
\end{gather}
where $\mathrm{proj}_c(\cdot)$ is the operator that projects a representation on the subspace of class $c$. The projection is illustrated in Fig. \ref{subspace}. Finally, we denote $\mathbf{d}_{i} = [d_{i, 1},...,d_{i, C}]^{\top} \in \mathbb{R}^C$.

\subsubsection{New Anomaly Detection}
The new type of anomaly should have large distances from the subspaces of all known classes. Based on the above principle, we can define the anomaly score $s_i$ for the $i\mathrm{th}$ sample as
\begin{equation}
    s_i = \min_c d_{i,c}.
\label{ano score}
\end{equation}
Notably, if $s_i$ exceeds a predefined threshold, the $i\mathrm{th}$ sample is identified as an anomaly.

\subsubsection{Anomaly Classification}
For the $i\mathrm{th}$ sample, which has been identified as known classes with a low anomaly score $s_i$, its probability $p_{i, c}$ of belonging to the $c\mathrm{th}$ class can be defined through the softmax function as:
\begin{equation}
p_{i,c}=p( \mathbf{y}_{i,c} = 1 \mid \mathbf{f}_i)=\frac{\exp \left(-d_{i,c}\right)}{\sum_{c^{\prime}} \exp \left(-d_{i,c^{\prime}}\right)}.
\label{prob} 
\end{equation}

According to the probabilities, the label of the input sample can be predicted. Finally, we denote $\mathbf{p}_{i} = [p_{i, 1},...,p_{i, C}]^{\top} \in \mathbb{R}^C$ for better illustration.

\subsection{Loss Function}

This subsection details the loss function design for learning effective representations and subspaces.

For anomaly classification, we first adopt the popular cross-entropy loss, which is defined as:
\begin{equation}
   \mathcal{L}_{\mathrm{ce}} = -\sum_i \mathbf{y}_{i}^{\top} \log \mathbf{p}_{i},
\end{equation}
where the $\log$ function operates on each entry of $\mathbf{p}_{i}$.

A crucial step is to ensure that we can learn valuable subspaces, i.e., each subspace is capable of accommodating its corresponding representations, while different subspaces maintain distinct separability during training. However, cross-entropy loss $\mathcal{L}_{\mathrm{ce}}$ is insufficient for our purpose. Therefore, to further strengthen our subspace properties, we propose another subspace deviation loss $\mathcal{L}_{\mathrm{dev}}$ that consists of the following parts, i.e., $\mathcal{L}_{\mathrm{dev}} = \mathcal{L}_{\mathrm{pos}} + \mathcal{L}_{\mathrm{neg}} + \mathcal{L}_{\mathrm{ineq}}$.

Denoted $\bar{d}_i$ as the distance of the $i\mathrm{th}$ sample to its associated subspace, while  $\tilde{d}_i$ as the minimum distance to any other subspace, we have
\begin{equation}
    \bar{d}_i = \{d_{i,c} | y_{i,c}=1\}, \tilde{d}_i = \min_c \{ d_{i,c} | y_{i,c}=0 \}.
\end{equation}

Specifically, $\mathcal{L}_{\mathrm{pos}}$ encourages the representations to fall into their corresponding subspaces:
\begin{equation}
\mathcal{L}_\mathrm{pos}=\max\{\max_i ( \bar{d}_i )-m_{\mathrm{n}}, 0\},
\end{equation}
 Notably, instead of enforcing a representation exactly on the subspace, we allow a small tolerance $m_{\mathrm{n}}$ to enable smoother optimization \citep{hadsell2006dimensionality}.

Furthermore, $\mathcal{L}_{\mathrm{neg}}$ aims to enhance the separation of subspaces, thereby augmenting the discriminative capacity for different classes. We formulate $\mathcal{L}_\mathrm{neg}$ as:
\begin{equation}
\mathcal{L}_\mathrm{neg}=\max\{m_{\mathrm{f}}-\min_i(\tilde{d}_i), 0\}.
\label{loss neg}
\end{equation}
The above Eq. \eqref{loss neg} indicates that the distances from the irrelevant subspaces are encouraged to be larger than $m_{\mathrm{f}}$.

Additionally, we propose $\mathcal{L}_{\mathrm{ineq}}$ to explicitly increase the difference between $\bar{d}_i$ and $\tilde{d}_i$:
\begin{equation}
\mathcal{L}_\mathrm{ineq}=\max\{\max_i(\bar{d}_i)-\min_i(\tilde{d}_i), 0\}.
\label{loss greater}
\end{equation}

Finally, the overall loss function $\mathcal{L}$ is as follows:
\begin{equation}
\begin{aligned}
\mathcal{L}=\mathcal{L}_{\mathrm{ce}}+\lambda \mathcal{L}_{\mathrm{dev}}
= \mathcal{L}_{\mathrm{ce}}+\lambda (\mathcal{L}_{\mathrm{pos}} + \mathcal{L}_{\mathrm{neg}}+\mathcal{L}_\mathrm{ineq}),
\label{loss}
\end{aligned}
\end{equation}
where $\lambda$ is hyperparameter to balance $\mathcal{L}_{\mathrm{ce}}$ and $\mathcal{L}_{\mathrm{dev}}$.

\subsection{Model Training}
\label{model training}

Accurate learning of the subspace parameters in our model is crucial. However, empirical observations indicate that random initialization of the subspaces is inefficient, as evidenced by representations failing to converge to the subspaces during training. The Singular Value Decomposition (SVD) provides an optimal low-rank approximation, making it suitable for initializing the subspaces in our scenarios.

We can collect all $m_c$ representations of the $c\mathrm{th}$ class in the training dataset as $\{\mathbf{f}_{i_c}, i_c=1,...,m_c\}$. Then, the SVD defines the center of $\{\mathbf{f}_{i_c}\}$ as the bias $\mathbf{b}_c$ of subspace:
\begin{equation}
    \mathbf{b}_c= \dfrac{1}{m_c}\sum_{i_c=1}^{m_c}\mathbf{f}_{i_c}.
\end{equation}
Removing the center  $\mathbf{b}_c$, we stack $\{\mathbf{f}_{i_c}-\mathbf{b}_c\}$ into a matrix $\mathbf{F}_c \in \mathbb{R}^{m_c\times p}$.  $\mathbf{F}_c$ can be decomposed by SVD as:
\begin{equation}
\mathbf{F}_c = \mathbf{U}_c \mathbf{\Lambda}_c \mathbf{V}_c^{\top},
\end{equation}
where $\mathbf{U}_c \in \mathbb{R}^{m_c \times m_c}$, $\mathbf{\Lambda}_c \in \mathbb{R}^{m_c \times p}$ is a non-negative diagonal matrix, and $\mathbf{V}_c=\left[\mathbf{v}_{1,c}, \cdots, \mathbf{v}_{p,c}\right] \in \mathbb{R}^{p \times p}$. Since $\mathbf{V}_c$ captures the principle directions, we can take the first $k$ column vectors $\mathbf{W}_c = \left[\mathbf{v}_{1,c}, \cdots, \mathbf{v}_{k,c} \right] \in \mathbb{R}^{p\times k}$ of matrix $\mathbf{V}$ as the initial bases of the subspace. From this perspective, the above SVD-based strategy minimizes the distances of representations to the initial subspace.

We now discuss to train our model in an end-to-end manner. We update the point encoder $\mathcal{E}_{\boldsymbol{\theta}}(\cdot)$ and subspaces $\left\{\mathbf{W}_c, \boldsymbol{b}_c\right\}$ by minimizing a mini-batch form of $\mathcal{L}$ in Eq. \eqref{loss}. During each of the first $N_{\mathrm{init}}$ iterations, we initialize the subspaces following the above SVD-based strategy. This two-layer loop procedure stops upon convergence. The training process is summarized in Algorithm \ref{alg}.

\begin{algorithm}
\caption{Model training and inference.}\label{alg}
\begin{algorithmic}[1]
\renewcommand{\algorithmicrequire}{\textbf{Stage 1:}}
\REQUIRE \textbf{Model training}
\renewcommand{\algorithmicrequire}{\textbf{Input:}}
\REQUIRE $ \left \{\mathbf{X}_i, \mathbf{y}_i\right \}$, $N_{init}$, $k$, $\lambda$, $m_{\mathrm{n}}$, and $m_{\mathrm{f}}$
\renewcommand{\algorithmicensure}{\textbf{Output:}}
\ENSURE $\mathcal{E}_{\boldsymbol{\theta}}$, $\left\{\mathbf{W}_c, \boldsymbol{b}_c\right\}$
\State Randomly initialize the model (including the point encoder and subspace classifier).
\State $N_{iter}=0$  
\REPEAT
    \IF{$N_{iter}<N_{init}$}
    \State Compute all representations $\{\mathbf{f}_i\}$ by Eq. \eqref{encoder}.
    \State Calculate the subspaces with $\left\{\mathbf{W}_c, \boldsymbol{b}_c\right\}$ by using SVD, as discussed in Section \ref{model training}.
    \ENDIF
    \State $N_{iter}=N_{iter}+1$ 
    \FOR{batch in $ \left \{\mathbf{X}_i\right \}$} 
    \State Calculate the mini-batch overall loss $\mathcal{L}$ in Eq. \eqref{loss}.
    \State Update the encoder $\mathcal{E}_{\boldsymbol{\theta}}$ and subspace classifier $\left\{(\mathbf{W}_c, \boldsymbol{b}_c)\right\}$.
    \ENDFOR
\UNTIL{converge}
\renewcommand{\algorithmicrequire}{\textbf{Stage 2:}}
\REQUIRE \textbf{Model inference}
\renewcommand{\algorithmicrequire}{\textbf{Input:}}
\REQUIRE $ \left \{\mathbf{X}_i\right \}$
\renewcommand{\algorithmicensure}{\textbf{Output:}}
\ENSURE $s_i$, $\mathbf{p}_{i}$
\State Load model parameters $\mathcal{E}_{\boldsymbol{\theta}}$, $\left\{(\mathbf{W}_c, \boldsymbol{b}_c)\right\}$. 
\State Calculate the representation $\mathbf{f}_i$ and deviation $\mathbf{d}_{i}$ by Eqs. \eqref{encoder} and \eqref{deviation cal}.
\State Calculate anomaly score $s_i$ by Eq. \eqref{ano score}, and identify new type of anomaly.
\State Compute the probabilities $\mathbf{p}_{i}$ for each class by Eq. \eqref{prob}.
\end{algorithmic}
\end{algorithm}

\section{Numerical Case Study}
\label{exp_num}

This section introduces a synthetic anomaly classification dataset to validate and compare our proposed methodology with state-of-the-art benchmarks comprehensively.

\subsection{Dataset Description}

\begin{table}[]
\tbl{Descriptions of our synthetic dataset.}
{\begin{tabular}{@{}ccccc@{}}
\toprule
                       & Class            & Sample Size & Standard model     & Parameter            \\ \midrule
\multirow{3}{*}{Known} & \textit{Dent}    & 100         & Half-ellipsoid     & Length; width; depth \\
                       & \textit{Scratch} & 100         & Half-cylinder      & Length; radius       \\
                       & \textit{Hole}    & 100         & Cone               & Height; radius       \\ \midrule
New type                & \textit{Groove}           & 30          & Long, narrow plane & Width; depth         \\ \bottomrule
\end{tabular}}
\label{anomaly generation}
\end{table}

The current public 3D anomaly classification datasets are limited and inadequate in representing the complexities of real manufacturing processes, such as intra-class variation, and inter-class similarity of anomalies, and do not support the task of new type of anomaly detection. Therefore, to effectively validate our proposed model, we introduce a new dataset where synthetic surface anomalies are designed in accordance with the definitions established by international standards, specifically ISO 8785 \citep{iso19998785}.

\subsubsection{Anomaly Generation}
We create 4 classes of anomalies in our dataset, including \textit{Dent}, \textit{Scratch},  \textit{Hole}, and \textit{Groove}. 

\begin{figure}[]
    \centerline{\includegraphics[width = 0.75\linewidth]{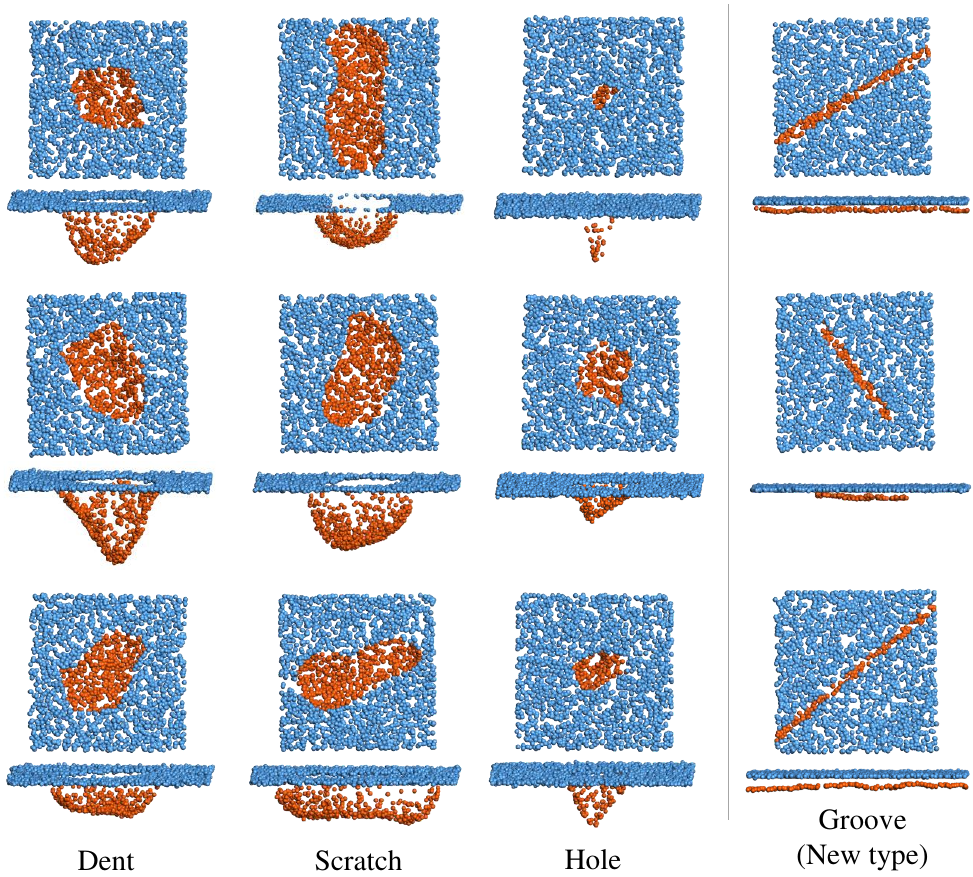}}
    \caption{The visualization of synthetic data. There are 4 classes of anomalies, including 3 types of known anomalies, i.e., \textit{Dent, Scratch, Hole}, and one new type of anomaly \textit{Groove}. Anomalies in this dataset exhibit intra-class variation and inter-class similarity.}
    \label{visualization}
\end{figure}

Following ISO 8785, we create standard models, such as ellipsoids and cones, for each anomaly class, with adjustable parameters to enable variations in size and shape. To enhance the variability of anomaly patterns that better reflect realistic scenarios, we apply non-rigid transformations using the direct average method with Gaussian weights \citep{arsigny2005polyrigid}. 
Additionally, random rigid transformations are applied to introduce diverse spatial poses.  A flat surface is also incorporated to represent the normal surface beyond anomalies. We visualize exemplars of each class in Fig. \ref{visualization}. Each class's standard model and adjustable parameters are detailed in Table \ref{anomaly generation}.

As many manufacturing applications may not have sufficient anomalous samples, 100 samples for each known class (\textit{Dent}, \textit{Scratch}, and \textit{Hole}) are generated for anomaly classification, divided into 30 samples for training, 10 for validation, and 60 for test dataset, respectively. Notably, we design a larger test dataset to comprehensively evaluate the model's generalization capability.
Additionally, we synthesize 30 \textit{Groove} samples for the new type of anomaly detection.

\subsection{Experimental Settings}

\subsubsection{Benchmarks}
\label{subsec: benchmarks}
For anomaly classification, we compare our deep subspace model with three popular architectures, including PointNet \citep{qi2017pointnet}, DGCNN \citep{wang2019dynamic}, and Point Cloud Transformer (PCT) \citep{guo2021pct}.
For the detection of a new type of anomaly, we compare with Support Vector Data Description (SVDD) \citep{tax2004support}, which trains an SVDD for each known class using the representations extracted from our point encoder. The final anomaly score is determined as the minimum from these SVDDs. The reason for selecting these comparison methods is that they represent the most widely used and currently state-of-the-art backbones in the field of anomaly classification, and have been proven to be highly effective.

\subsubsection{Evaluation Metrics}
We evaluate the classification performance using accuracy (ACC), balanced accuracy (BA), Recall, Precision, and F1-score. In addition, we utilize the area under the receiver operating characteristic (AUROC) for new types of anomaly detection. 

\textbf{Accuracy} measures the overall correctness of the classification model, indicating the proportion of correctly classified instances out of all instances.
\begin{equation}
\text{ACC} = \frac{\text{TP} + \text{TN}}{\text{TP} + \text{TN} + \text{FP} + \text{FN}}.
\end{equation}

\textbf{Balanced Accuracy} balances the accuracy on both the positive and negative classes, making it suitable for imbalanced datasets where accuracy might be misleading.
\begin{equation}
\text{BA} = \frac{\text{TPR} + \text{TNR}}{2},
\end{equation}
where
\begin{equation}
\text{TPR} = \frac{\text{TP}}{\text{TP} + \text{FN}} \quad \text{and} \quad \text{TNR} = \frac{\text{TN}}{\text{TN} + \text{FP}}.
\end{equation}

\textbf{Recall} evaluates the model's ability to find all relevant instances, i.e., the proportion of actual positives that are correctly identified.
\begin{equation}
\text{Recall} = \frac{\text{TP}}{\text{TP} + \text{FN}}.
\end{equation}

\textbf{Precision} measures the accuracy of the positive predictions, indicating the proportion of predicted positives that are actually positive.
\begin{equation}
\text{Precision} = \frac{\text{TP}}{\text{TP} + \text{FP}}.
\end{equation}

\textbf{F1-score} provides a balance between precision and recall, useful for scenarios where both are important.
\begin{equation}
\text{F1} = 2 \times \frac{\text{Precision} \times \text{Recall}}{\text{Precision} + \text{Recall}}.
\end{equation}

\textbf{Area Under the Receiver Operating Characteristic (AUROC)}. The AUROC measures the model's ability to detect new type of anomaly, with a higher AUROC indicating better performance in anomaly detection. The AUROC is calculated as the area under the curve of the True Positive Rate (TPR) plotted against the False Positive Rate (FPR):
\begin{equation}
\text{AUROC} = \int_{0}^{1} \text{TPR}(FPR) \, dFPR,
\end{equation}
where 
\begin{equation}
\text{FPR} = \frac{\text{FP}}{\text{FP} + \text{TN}}.
\end{equation}

\subsubsection{Statistical Testing}
\label{sign test}
For classification, we conduct 30 replications by randomly splitting the original dataset,  following \citep{zhang2022contrastive,zhang2021path}. As the results may not follow Gaussian distributions strictly, we employ the sign test \citep{demvsar2006statistical} for the significance test. The null hypothesis ($H_0$) posits that the performance of the proposed methodology is not superior to benchmark methodologies. At the significance level of 5\%, the indicator variable ($H$) equals 1 if $H_0$ is rejected. In contrast, if we accept $H_0$, $H=0$.

\subsubsection{Implementation}
During the training phase, our model is optimized using the Adam algorithm, configured with an initial learning rate of 0.001, weight decay set to 0.0001, and a batch size of 32. The learning rate undergoes adjustment via the cosine annealing strategy. Moreover, we stop the processes of subspace model training when reaching 150 epochs.

For the hyperparameters, we set the dimension of subspace $k=20$, tuning parameter $\lambda=1$, $N_{init}=30$, the margins in $\mathcal{L}_{\mathrm{dev}}$ $m_{\text{n}}=0$ and $m_{\text{f}}=1$. All the experiments were run on an NVIDIA RTX A6000 GPU.

\subsection{Anomaly Classification Results}
\label{syn exp}

\begin{figure}[]
    \centerline{\includegraphics[width = 0.7\linewidth]{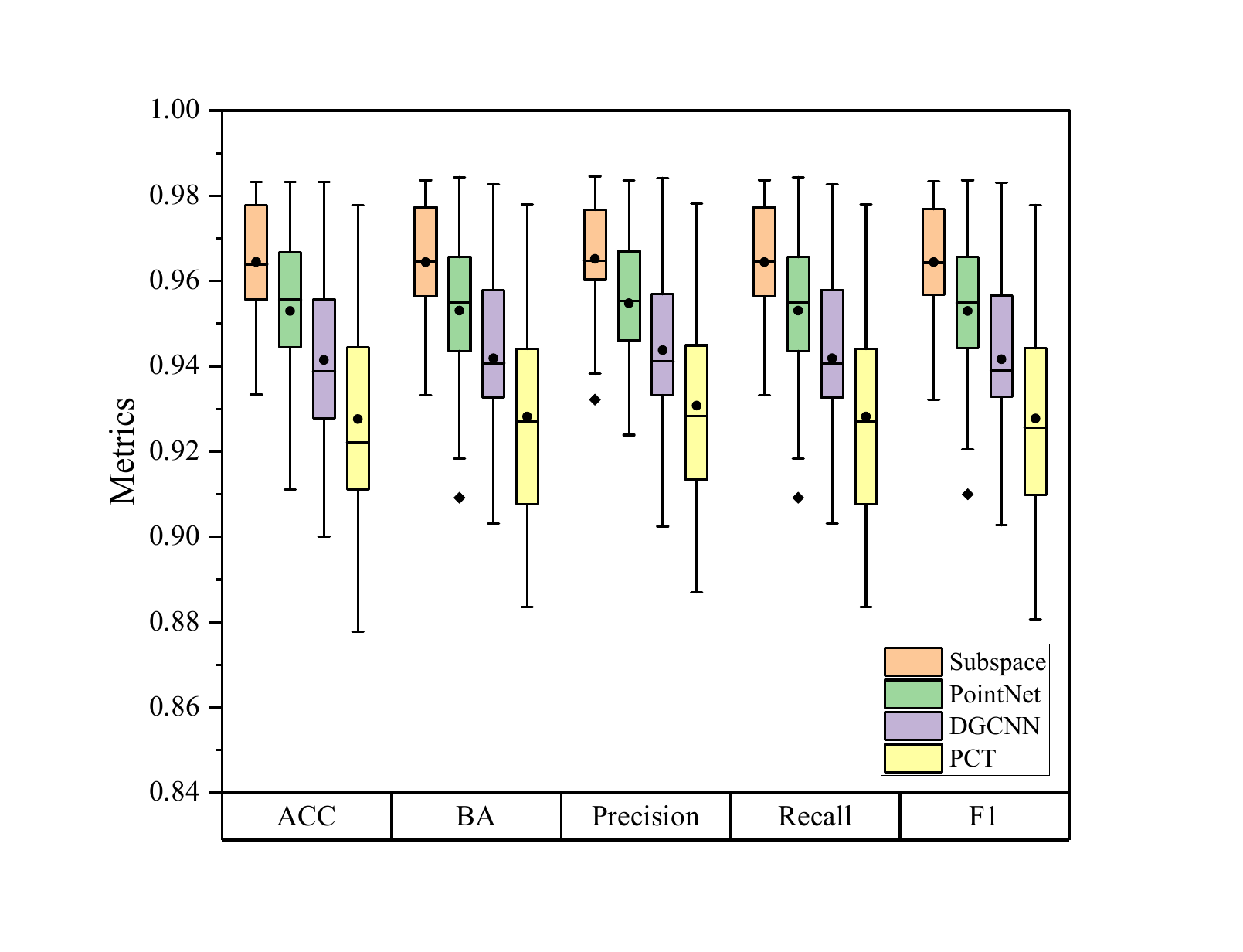}}
    \caption{The boxplots of different methods in terms of classification metrics.}
    \label{syn_boxplot}
\end{figure}

Table \ref{syn cls} and Fig. \ref{syn_boxplot} show that our model achieves
the highest evaluation metrics and lowest standard deviations. The statistical test in Table \ref{sign test:syn} shows that the p-values are less than 0.05 for all evaluation metrics, claiming that our model significantly outperforms other benchmarks.

\begin{table}[h]
\tbl{The classification results on synthetic dataset.}
{\begin{tabular}{@{}cccccc@{}}
\toprule
Methods  & ACC      & BA  & Precision     & Recall        & F1-score      \\ \midrule
PointNet & 0.9529  & 0.9530  & 0.9547  & 0.9530  & 0.9529  \\
DGCNN    & 0.9415  & 0.9419  & 0.9438  & 0.9419  & 0.9416  \\
PCT      & 0.9276  & 0.9282  & 0.9308  & 0.9282  & 0.9277  \\
Subspace & \textbf{0.9644 } & \textbf{0.9644 } & \textbf{0.9652 } & \textbf{0.9644 } & \textbf{0.9644 } \\ \bottomrule
\end{tabular}}
\label{syn cls}
\end{table}

\begin{table}[h]
\tbl{The significance test results (p-values) on synthetic dataset.}
{\begin{tabular}{@{}cccc@{}}
\toprule
Synthetic & PointNet & DGCNN   & PCT      \\ \midrule
ACC       & \num{2.97e-5}  & \num{1.60e-4} & \num{9.31e-10} \\
BA        & \num{7.10e-4}  & \num{2.60e-3} & \num{9.31e-10} \\
Precision & \num{8.00e-3}  & \num{2.60e-3} & \num{9.31e-10} \\
Recall    & \num{7.10e-4}  & \num{2.60e-3} & \num{9.31e-10} \\
F1-score  & \num{7.10e-4}  & \num{7.10e-4} & \num{9.31e-10} \\ \bottomrule
\end{tabular}}
\label{sign test:syn}
\end{table}

\begin{figure}[]
    \centerline{\includegraphics[width = \linewidth]{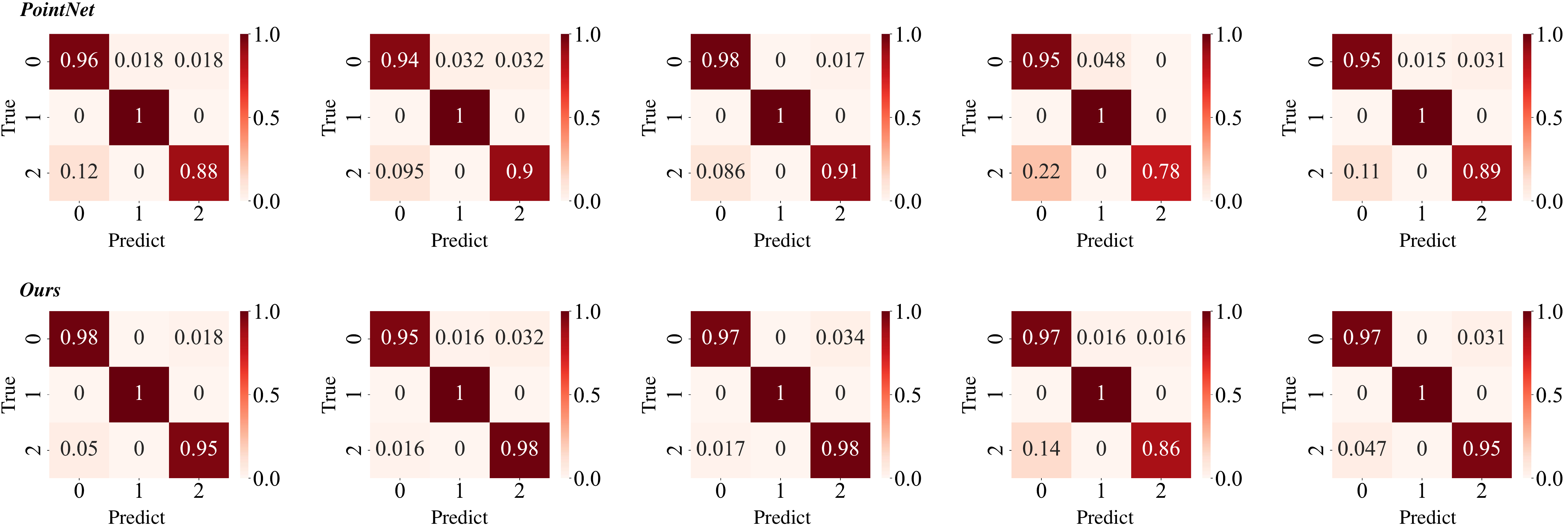}}
    \caption{The confusion matrices of classification results for five random replications. The upper row displays baseline (PointNet) results, while the bottom row shows the results of our subspace model. (The labels 0, 1, and 2 represent \textit{Dent}, \textit{Hole}, and \textit{Scratch}, respectively.)}
    \label{cls matrix}
\end{figure}

With more complex network structures, DGCNN and PCT tend to overfit the training data, decreasing their effectiveness. While our model's point encoder is similar to that of PointNet, we achieve better results due to our proposed subspace classifier, which effectively addresses the challenges of our task.
Moreover, Fig. \ref{cls matrix} presents the confusion matrices for the proposed model (bottom row) and the baseline PointNet (upper row) in five replications. We observe that the classification of \textit{Dent} and \textit{Scratch} is prone to errors, with baseline methods achieving an accuracy of approximately 0.9. This is primarily due to the intra-class variation and inter-class similarity present in these anomaly types, as shown in Fig. \ref{visualization}. In contrast, our proposed method achieves classification accuracies of around 0.95 for these categories. This improvement indicates that the subspace effectively addresses the challenges posed by intra-class variation and inter-class similarity, thereby enhancing overall classification performance.

\begin{figure}[]
    \centerline{\includegraphics[width = 0.95\linewidth]{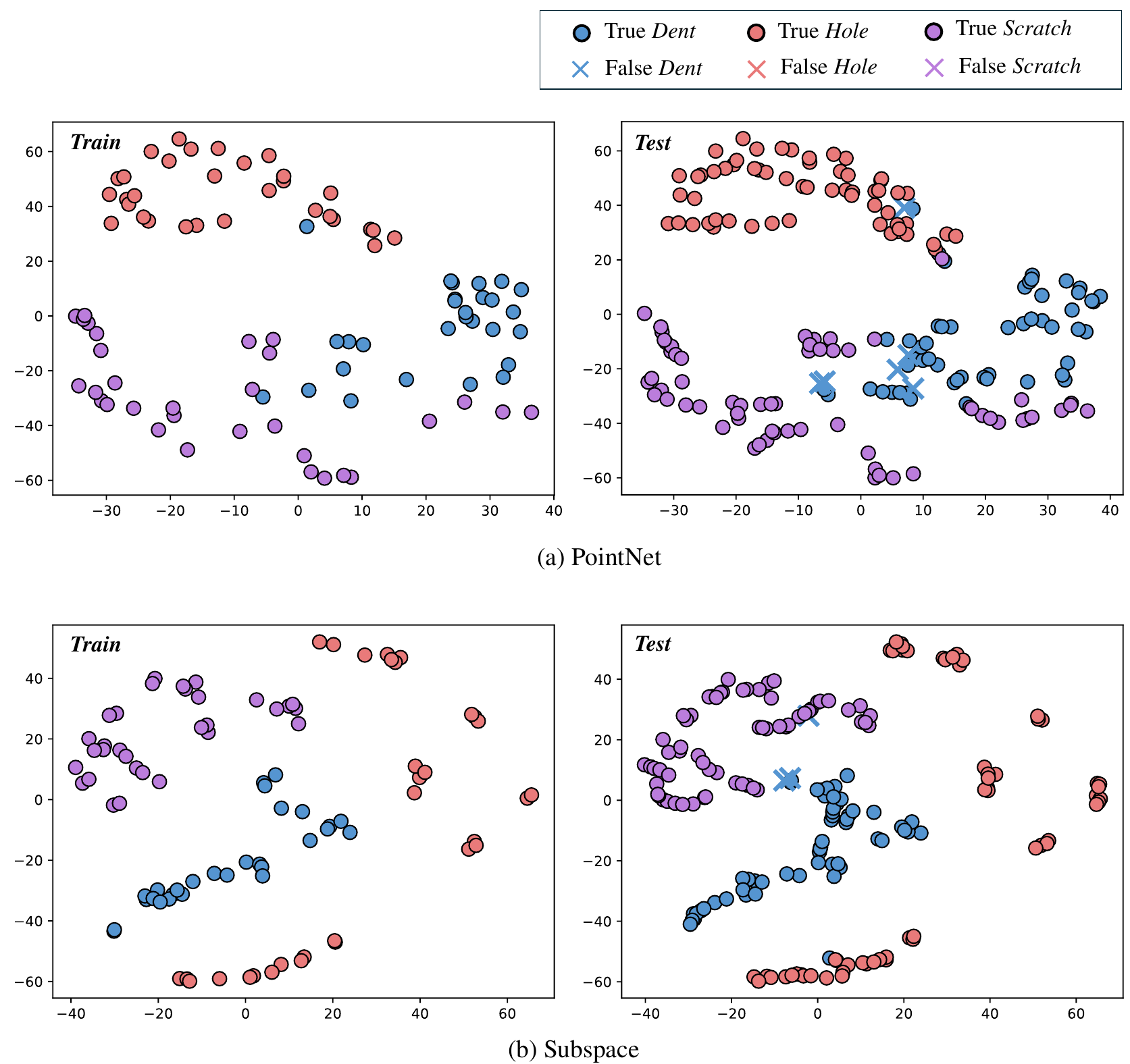}}
    \caption{The representation distributions of training and test datasets processed by tSNE in (a) PointNet and (b) our subspace model.}
    \label{feat distribution}
\end{figure}

\begin{figure}[]
    \centerline{\includegraphics[width =\linewidth]{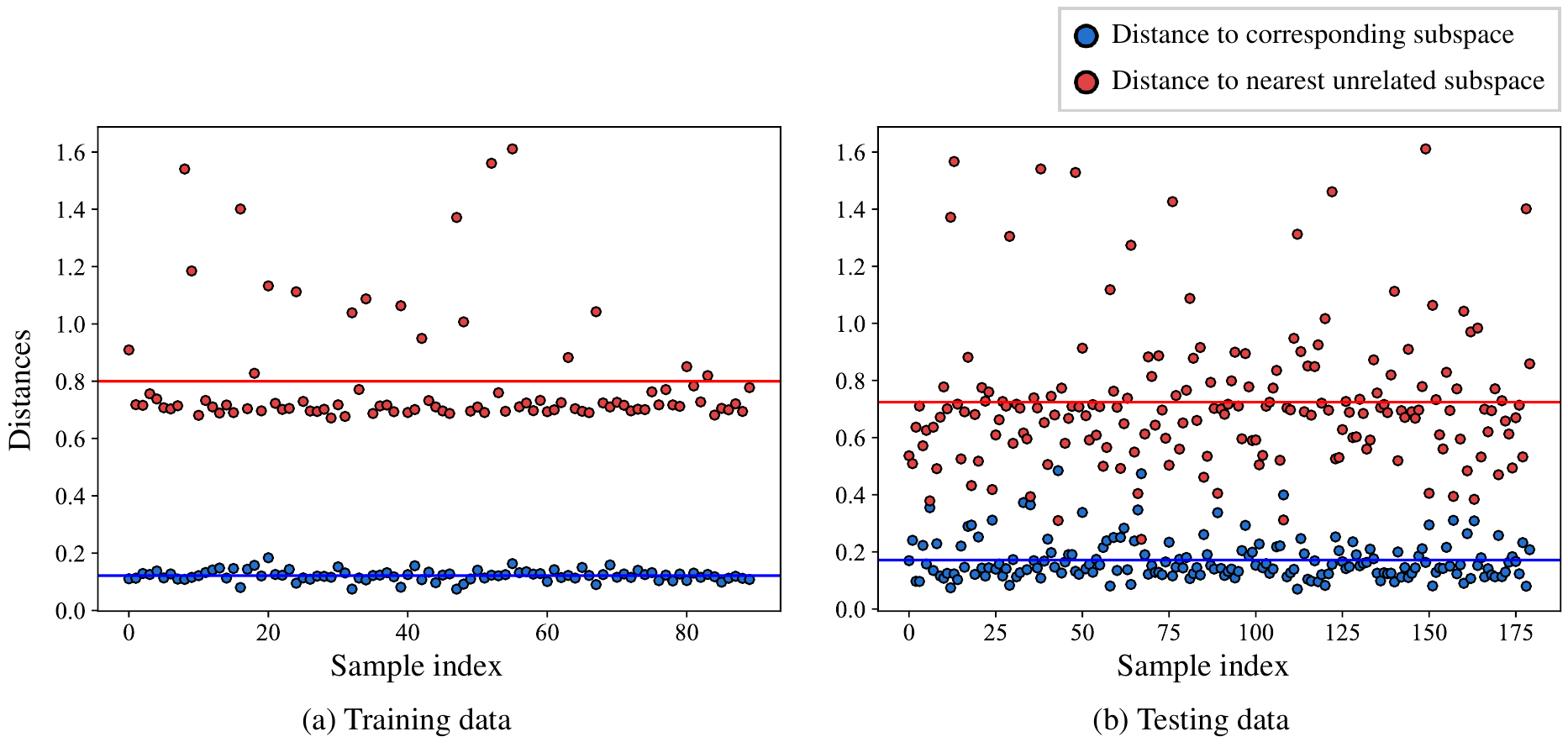}}
        \caption{The distances of representations to the corresponding (blue dots) and nearest unrelated subspace (red dots). The solid lines depict the mean distance values for each class.}
    \label{subspace dists}
\end{figure}

\subsection{Representation Visualization}
\label{feat exp}

To interpret the effectiveness of our model, we first visualize the learned representations from our model and PointNet in the training and test datasets by t-SNE in Fig. \ref{feat distribution}. Specifically, our model demonstrates a more distinct separation between features from different classes.
Moreover, the distributions of training and testing features shown in Fig. \ref{feat distribution}(b) are nearly identical, demonstrating the good generalization ability of our model.

Fig. \ref{subspace dists} shows the distance of each sample to its corresponding subspace (blue dots) and the nearest unrelated subspace (red dots). The distances to the corresponding subspaces are close to zero, indicating that the learned subspaces accurately model the representation distributions. Additionally, the distances to the nearest unrelated subspaces are adequately large, highlighting the subspaces' discriminative power for anomaly classification.

\subsection{Sensitivity Analysis}

\begin{figure}[]
    \centerline{\includegraphics[width = \linewidth]{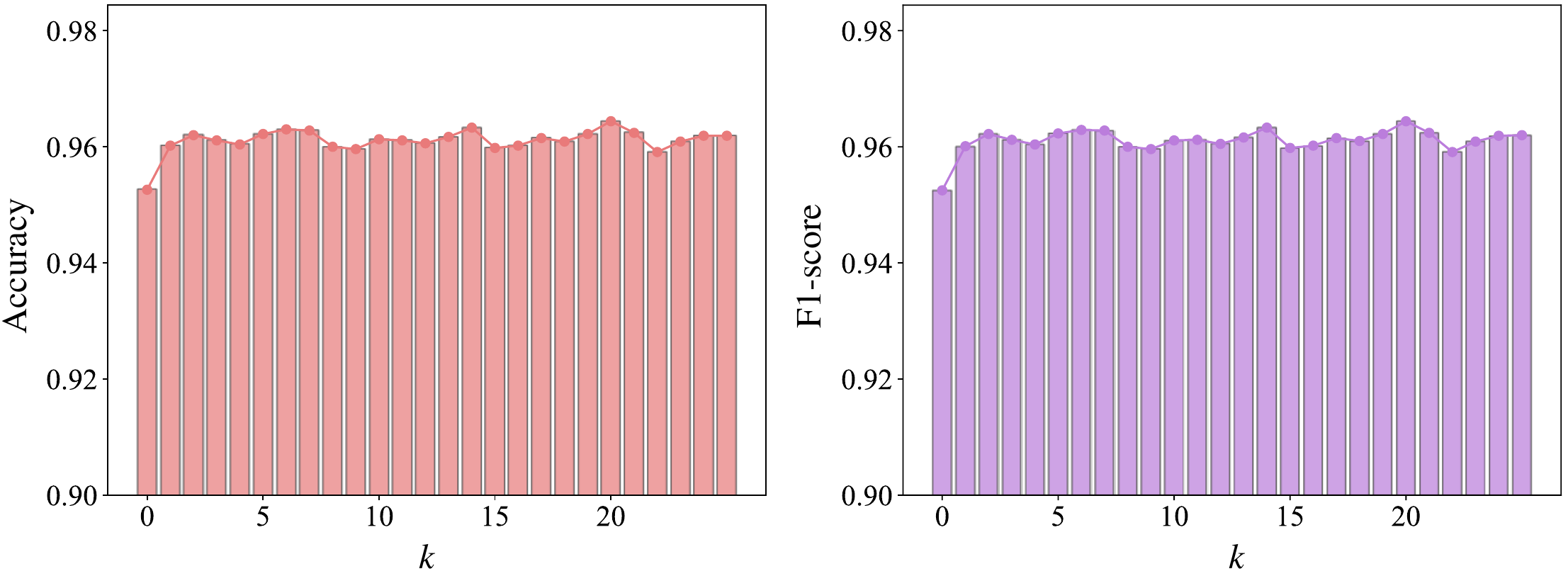}}
    \caption{Sensitivity analysis under different subspace dimension $k$.}
    \label{sensitivity}
\end{figure}

In this section, we perform a sensitivity analysis of the subspace dimension $k$ on the classification performance within the range $k \in [0, 25]$. Fig. \ref{sensitivity} summarizes the F1-scores under different $k$. Our model is relatively stable across different $k$ around 0.96, while decreases to 0.9526 at $k=0$. The reason is that the subspace degenerates into a singleton. Therefore, the model pulls representations from the same class close together while pushing those from different classes farther apart. This mechanism prevents the modeling of intra-class variation, thus influencing the classification performance. Notably, we achieve the best performance at $k=20$, which is also selected for our experiments on synthetic data.

\subsection{New Type of Anomaly Identification}

\begin{table}[h]
\tbl{The results of new type of anomaly identification on synthetic dataset (mean $\pm$ std).}
{\begin{tabular}{@{}ccc@{}}
\toprule
Methods & Our encoder + Subspace   & Our encoder + SVDD   \\ \midrule
AUROC   & \textbf{0.9980 $\pm$ 0.003} & 0.8945 $\pm$ 0.051\\ \bottomrule
\end{tabular}}
\label{ano detection}
\end{table}

In this subsection, we assess the efficacy of our subspace model in identifying the new type (\textit{Groove}) of  anomaly. We compare with the SVDD method as described in Section \ref{subsec: benchmarks}. The results of AUROC are summarized in Table \ref{ano detection}.

Table \ref{ano detection} demonstrates that our model yields promising outcomes, achieving an average AUROC of 0.9980 with a very low standard deviation of 0.003 across 30 replications. The superior performance of our model is attributed to its capacity to effectively learn the distributions of known classes as subspaces while ensuring that new types of anomalies remain significantly distant from these subspaces.

\section{Real Case Study}
\label{exp_case}

\begin{figure}[]
    \centerline{\includegraphics[width = \linewidth]{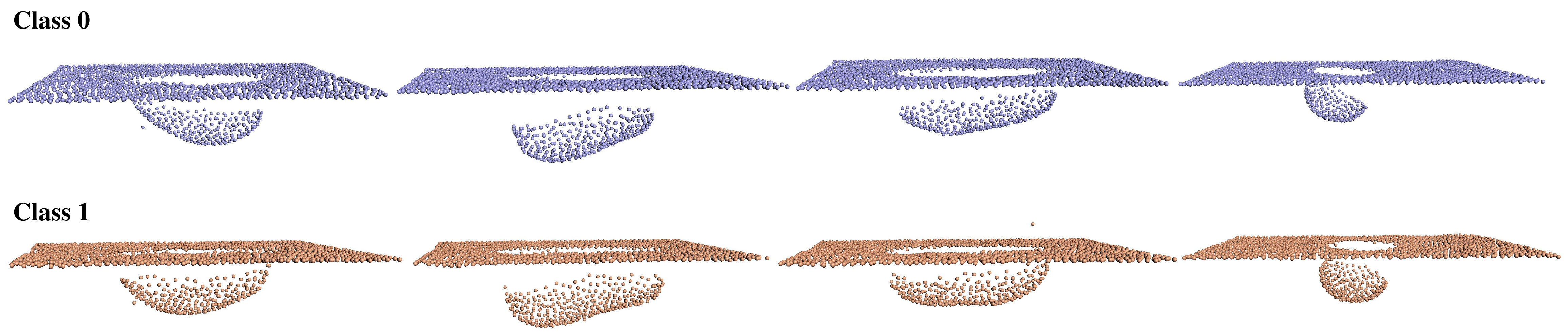}}
    \caption{The visualization of real scanned data. Different rows represent different classes, and samples within the same row belong to the same class. This dataset has a obvious intra-class variation and inter-class similarity property.}
    \label{real data}
\end{figure}

In this section, we validate our method using a real dataset from the 2023 IISE data analytics competition\footnote{https://qaweb.iise.org/Details.aspx?id=50000}. This dataset contains two classes of anomalies of a manufactured product, where there are 190 samples and each point cloud sample contains 2048 measurement points. As shown in Fig. \ref{real data}, the anomalies from the same class have various geometrical shapes and sizes, while some anomalies from different types can be similar.
Therefore,
this dataset exhibits strong intra-class variation and inter-class similarity. 

\begin{table}[h]
\tbl{The classification results on real dataset.}
{\begin{tabular}{@{}cccccc@{}}
\toprule
Methods  & ACC               & BA          & Precision              & Recall                 & F1-score               \\ \midrule
PointNet & 0.8651           & 0.8299           & 0.8621           & 0.8299           & 0.8397           \\
DGCNN    & 0.8018           & 0.7515           & 0.7893           & 0.7515           & 0.7587           \\
PCT      & 0.7485           & 0.7008           & 0.7214           & 0.7008           & 0.7023           \\
Subspace & \textbf{0.9076}  & \textbf{0.8919}  & \textbf{0.8994}  & \textbf{0.8919}  & \textbf{0.8931}  \\ \bottomrule
\end{tabular}}
\label{real cls}
\end{table}

\begin{figure}[]
    \centerline{\includegraphics[width = 0.7\linewidth]{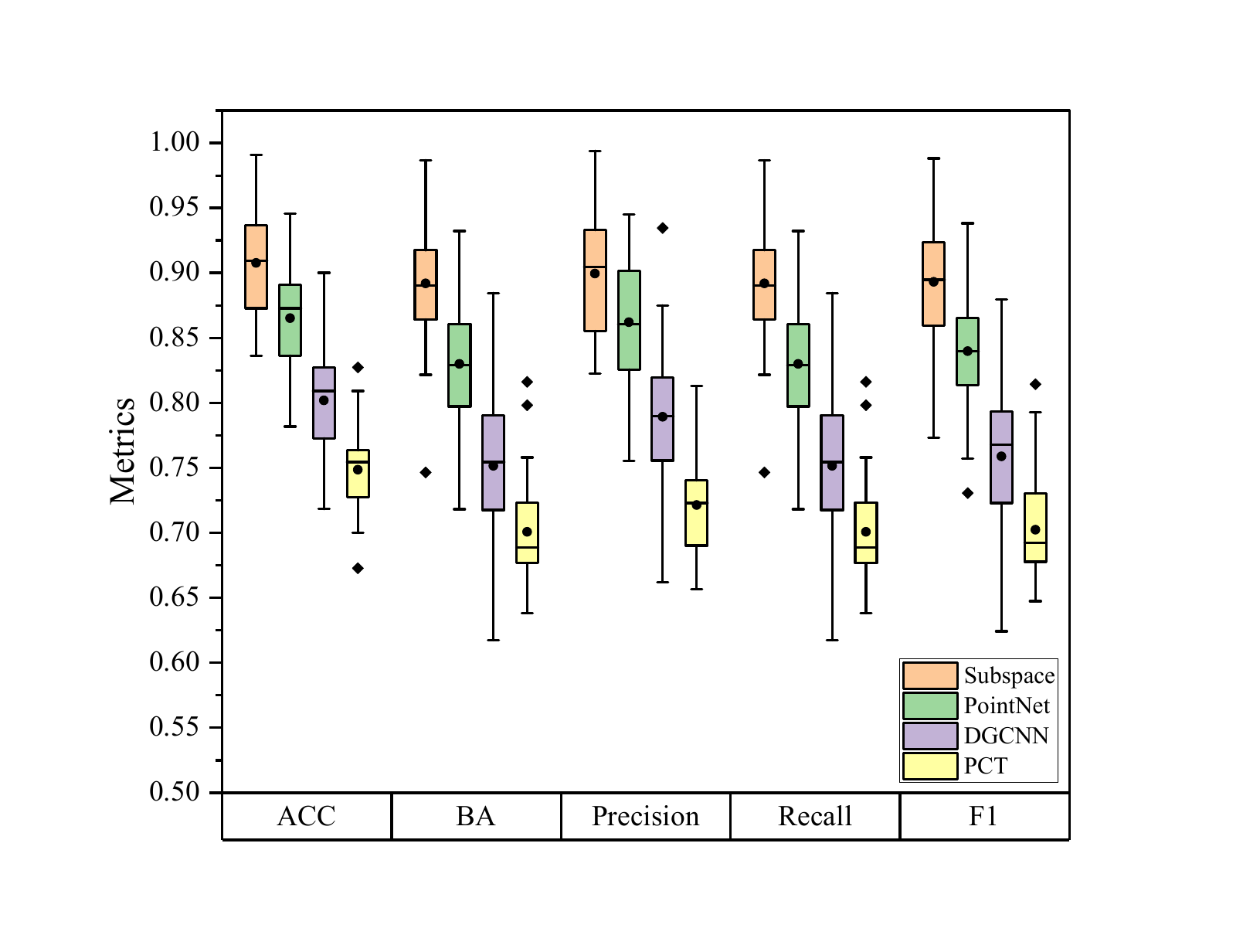}}
    \caption{The boxplots of different methods in terms of classification metrics on real dataset.}
    \label{real_boxplot}
\end{figure}

For the experimental setting, we split the original dataset into 60, 20, and 110 samples for train, validation, and test datasets, respectively. Notably, the large test dataset serves to comprehensively evaluate our model's generalization ability. Likewise, we conduct 30 replications on this dataset. The settings of the tuning parameters are: $k=12, \lambda=1, m_{\text{n}}=0, m_{\text{f}}=20$, $N_{init}=0$, and other parameters remain unchanged.

The results are described in Table \ref{real cls} and Fig. \ref{real_boxplot}. Our method advances significantly more than the benchmarks, with a BA of 0.8919. In contrast, the best benchmark, i.e., PointNet, only achieves a BA of 0.8299, while PCT and DGCNN show even much poorer results due to their complex network structures.
This enhancement is attributed to the more pronounced intra-class variation and inter-class similarity characteristics in real data, which further amplifies the advantages of our subspace model. 

Table \ref{sign test:real} presents the results of the sign test (conducted under the same experimental conditions as described in Section \ref{sign test}), with all metrics showing p-values below 0.05. This indicates that our method is statistically significantly superior to the benchmark methods.

\begin{table}[h]
\tbl{The significance test results (p-values) on real dataset.}
{\begin{tabular}{@{}cccc@{}}
\toprule
Real      & PointNet & DGCNN    & PCT      \\ \midrule
ACC       & \num{2.97e-5}  & \num{9.31e-10} & \num{9.31e-10} \\
BA        & \num{2.97e-5}  & \num{2.88e-8}  & \num{9.31e-10} \\
Precision & \num{8.00e-3}  & \num{9.31e-10} & \num{9.31e-10} \\
Recall    & \num{2.97e-5}  & \num{2.88e-8}  & \num{9.31e-10} \\
F1-score  & \num{2.97e-5}  & \num{2.88e-8}  & \num{9.31e-10} \\ \bottomrule
\end{tabular}}
\label{sign test:real}
\end{table}

In general, both synthetic and real case studies demonstrate that our subspace model achieves accurate anomaly classification and new type of anomaly identification simultaneously, outperforming other benchmark methods significantly.

\section{Conclusion}
\label{conclusion}
Anomaly classification plays an essential role in practical manufacturing systems for fault diagnosis and quality improvement. However, there is limited research focusing on the accurate classification of surface anomalies characterized by intra-class variation and inter-class similarity, as well as the simultaneous identification of new types of anomalies. In this paper, we proposed a novel deep subspace model to tackle the above issue. Specifically, we employed a lightweight point encoder to learn the representations of point cloud data, directly followed by a subspace classifier. Furthermore, we designed a novel loss function and a related training strategy to efficiently train our model. By modeling the distribution of each class in a sophisticated way, the well-trained subspace classifier effectively manages intra-class variation and inter-class similarity while delineating a precise boundary for known classes, thus being able to detect new types of anomalies in the test data. The comprehensive numerical studies on synthetic and real datasets demonstrated the efficacy of our proposed method.



\section*{Disclosure statement}

No potential conflict of interest was reported by the authors.

\section*{Data availability statement}

The data that support the findings of this study are available from the corresponding author upon reasonable request.

\section*{Funding}

This work was supported by the National Natural Science Foundation of China under Grant No. 72371219, No. 72001139 and No. 52372308, Guangdong Basic and Applied Basic Research Foundation under Grant No. 2023A1515011656, and Guangzhou-HKUST(GZ) Joint Funding Program under Grant No. 2023A03J0651 and No. 2024A03J0680.





\bibliographystyle{chicago}
\bibliography{ref}

\end{document}